# DeepSeaNet: Improving Underwater Object Detection using EfficientDet

Sanyam Jain
*Dept. of Computer Science and Communication,*
*Østfold University College*
Halden, Norway
sanyamj@hiof.no

*Abstract*— Marine animals and deep underwater objects are difficult to recognize and monitor for safety of aquatic life. There is an increasing challenge when the water is saline with granular particles and impurities. In such natural adversarial environment, traditional approaches like CNN start to fail and are expensive to compute. This project involves implementing and evaluating various object detection models, including EfficientDet, YOLOv5, YOLOv8, and Detectron2, on an existing annotated underwater dataset, called the "Brackish-Dataset". The dataset comprises annotated image sequences of fish, crabs, starfish, and other aquatic animals captured in Limfjorden water with limited visibility. The aim of this research project is to study the efficiency of newer models on the same dataset and contrast them with the previous results based on accuracy and inference time. Firstly, I compare the results of YOLOv3 (31.10% mean Average Precision (mAP)), YOLOv4 (83.72% mAP), YOLOv5 (97.6%), YOLOv8 (98.20%), EfficientDet (98.56% mAP) and Detectron2 (95.20% mAP) on the same dataset. Secondly, I provide a modified BiSkFPN mechanism (BiFPN neck with skip connections) to perform complex feature fusion in adversarial noise which makes modified EfficientDet robust to perturbations. Third, analyzed the effect on accuracy of EfficientDet (98.63% mAP) and YOLOv5 by adversarial learning (98.04% mAP). Last, I provide class-activation-map based explanations (CAM) for the two models to promote Explainability in black box models. Overall, the results indicate that modified EfficientDet achieved higher accuracy with five-fold cross validation than the other models with 88.54% IoU of feature maps.

## I. INTRODUCTION

Automated Maritime Object Detection and Marine Vision (AMODMV) [1] helps in scientific research and deep underwater wildlife exploration. It ensures safe navigation to avoid potential collisions with seabed objects, environmental protection by identifying and tracking pollution, preventing-preserving flora and fauna of aquatic life and security in international waters by detecting and tracking boats and ships that may be engaged in illegal activities like smuggling or piracy. This has much larger application usage for ensuring the safety of ships and crew, conducting search and rescue operations. Huge impact on aquatic life, due to pollution, overfishing, invasive species, and infrastructure intervention leads to loss of biodiversity, imbalance of food chain, and contamination of seafood. Therefore, requirement of continuous monitoring, controlled fishing, counting population, and identifying individual species is key to conserving and protecting our resources. A quick overview of my complete working pipeline is shown in Figure 1. Details are further discussed in the remaining paper.

While performing AMODMV, biggest challenge is the noisy images (limited visibility) because of the impurities of water, wrong placement angles of image sensing kits, and interpreting the results from the object detection model. In addition, The marine environment is often cluttered with a variety of objects, such as rocks, plants, and debris, which can make it difficult to distinguish between objects of interest and background clutter [2,3]. I experimented and proved the hypothesis, "Can we learn a noisy deep-sea imagery using a real world recorded video with the help of annotations using EfficientDet algorithm performed on the Brackish Dataset?" to be true. Not only statistical proves, but also provide saliency map and CAM based results that prove model has learned correct class specific features with respective confidence percentage. More advanced combinations and tricks are performed during the experiment using Adversarial Learning for higher accuracy (along with Precision and Recall). In the best of my knowledge, this work is the first attempt to improve the baseline of the Brackish Dataset [15].

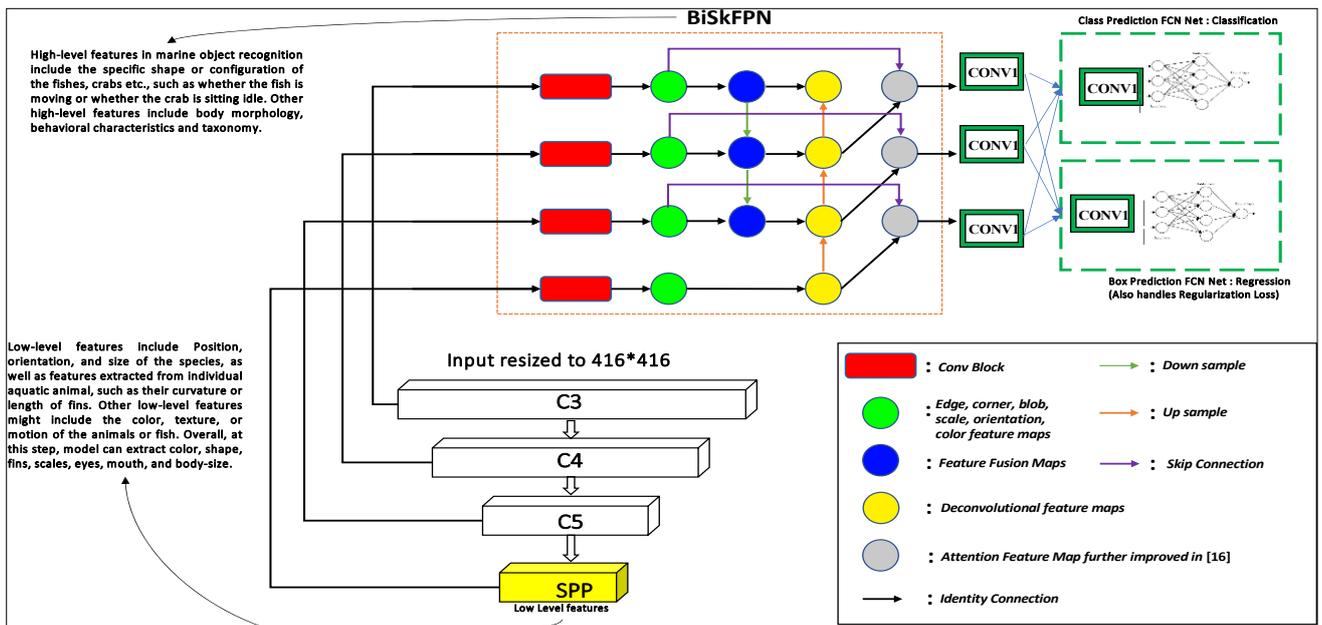

*Figure 1* An overview of proposed EfficientDet Object Detection Working Pipeline using EfficientNet backbone, BiSkFPN bottleneck and MFL head.

One Stage Object Detector (OSOD) and Two Stage Object Detectors (TSOD) have reportedly proven better than traditional approaches of Object Detection (e.g., sliding window method, SIFT, and HOG). OSOD models (for e.g., YOLO), can quickly and accurately detect objects in a single pass. They achieve this by predicting the class and location of objects using a single network architecture. However, one-stage models can struggle with accurately detecting small objects or objects with complex shapes. TSODs such as Faster R-CNN and Mask R-CNN, take a different approach. These models first generate a set of region proposals, which are likely locations for objects in the image. They then classify and refine these proposals to determine the final object detection. This approach allows for more accurate detection of small objects or objects with complex shapes but is generally slower than one-stage models [4,5].

Single-stage object detectors, such as YOLO (You Only Look Once) [6] and EfficientDet [7], have become increasingly popular in object detection tasks due to their speed, accuracy, and efficiency. YOLO uses a single-stage detection pipeline that consists of a backbone, neck, and head. The backbone is typically a feature extractor, such as DarkNet-53 or CSPDarkNet-53 [8], that extracts feature from the input image. The neck module, such as FPN [9] or PANet [10], is responsible for fusing features across different levels of the network, which helps to improve the detection performance of small and large objects. The head then performs the final detection task, which involves predicting bounding boxes and object classes directly from the feature maps.

EfficientDet is a variant of YOLO that uses an efficient backbone network and a BiFPN (Bidirectional Feature Pyramid Network) [7] to fuse features across different levels of the network. EfficientDet is known for balancing the speed vs accuracy tradeoff and cheap parameter space. The backbone is a compound scaling of EfficientNet [11] that consists of a series of efficient building blocks, such as MBConv (inverted bottleneck), that are optimized for a small computational budget. The modified BiSkFPN mechanism improves feature fusion in adversarial noise by using a skip connections and weighted average of the features in the top-down and bottom-up pathways. This makes EfficientDet more robust to perturbations and more accurate than other single-stage detectors. The BiSkFPN also includes skip connections that allow the network to use high-resolution feature maps for small objects and low-resolution feature maps for large objects, which improves the detection performance across a wide range of object sizes.

Two-stage object detection methods, such as Faster R-CNN [12] and Mask R-CNN [13], are not suitable for AMODMV due to their high computational requirements and slow inference time. On the other hand, single-stage object detection methods are faster and more efficient, making them better suited for this application. EfficientDet has shown to perform better than other single-stage object detection models such as YOLOv5 and Detectron2 [14] on the Brackish Dataset [15]. This is due to its efficient backbone network, BiSkFPN mechanism which allow it to effectively deal with the challenges posed by the marine environment.

## II. CONTRIBUTIONS

In this section, I present the contributions of my research project aimed at improving underwater object detection using state-of-the-art object detection models and techniques:

1. Comparative evaluation of state-of-the-art object detection models: The project involves a comparative evaluation of various object detection models, including EfficientDet, YOLOv5, YOLOv8, and Detectron2, on an existing annotated underwater dataset. This evaluation helps to identify the most effective model for underwater object detection.

2. Development of a modified BiSkFPN mechanism: The project proposes a modified BiSkFPN mechanism, which involves using a BiFPN neck with skip connections, to perform complex feature fusion in adversarial noise. This mechanism makes the modified EfficientDet model more robust to perturbations and improves its accuracy.

3. Adversarial learning for improved object detection accuracy: The project analyzes the effect of adversarial learning on the accuracy of EfficientDet and YOLOv5. This analysis helps to identify the most effective training strategy for underwater object detection.

4. Explainability in black box models: The project provides class-activation-map based explanations (CAM) for EfficientDet and YOLOv5 to promote Explainability in black box models. These explanations help to improve the interpretability of the models and enhance their trustworthiness.

5. Improved performance in AMODMV: The project demonstrates that modified EfficientDet achieved higher accuracy with five-fold cross-validation than other models. The use of EfficientDet improves the accuracy and reliability of AMODMV, which can enhance maritime security, monitor the environment, and protect aquatic life.

## III. RELATED WORK

This section presents summaries of my study on modified architectures specially for complex computer vision techniques and monitoring applications. Firstly, I review five recent works that has used EfficientDet algorithm along with other OSOD. Secondly, I study FPN, PANet and BiFPN bottlenecks used in OSOD algorithms. Finally, I study several concepts form the related research that will be used in later parts of the paper.

### A. 3.1-mynet: Improved EfficientDet using Attention Mechanism (AM) – Multiclass Focal Loss (MFL):

A new method that uses AM to dampen the effect of noise (caused by pollution, clouds, and climate) in remote sensing images. This work [16] also modifies pooling in every layer such that it can capture tiny class specific pixels and hence uses exhaustive feature space. This approach increases computational complexity but helps to achieve higher accuracy. It is because of the residual deformable 3-D convolution (RD3C) which extends the traditional 2-D convolution operation to better capture object deformations and variations in 3-D data (for e.g., space imagery or remote sensing). Two basic operations that are used in the work are 3D-Convolutional operation and Geo-Spatial Deformable-3D Convolutional Operation which is further explained in the following equations. The standard 3D convolution operation can be represented as:

$$y_{i,j,k} = \sum_{d=-D}^{D} \sum_{h=-H}^{H} \sum_{w=-W}^{W} x_{i+d,j+h,k+w} \cdot k_{d,h,w}$$

where $x$ is the input volume, $y$ is the output volume, $k$ is the convolution kernel, and $D$, $H$, and $W$ are the depth, height, and width of the kernel, respectively. In RD3C, the 3D deformable convolution operation is introduced before the standard 3D convolution, which can be represented as:

$$z_{i,j,k} = \sum_{d=-D}^{D} \sum_{h=-H}^{H} \sum_{w=-W}^{W} x_{i+d,j+h,k+w} \cdot k_{d,h,w} + \sum_{p=1}^{P} \sum_{q=1}^{Q} \sum_{r=1}^{R} x_{i+p,j+q,k+r} \cdot \Delta k_{p,q,r}$$

where $z$ is the intermediate feature map obtained by the deformable convolution operation, $P$, $Q$, and $R$ are the depth, height, and width of the offset kernel, respectively, and $\Delta k$ is the learnable deformation offset applied to the kernel. The deformation offset is learned from the input features using a separate convolutional operation, which can be represented as:

$$\Delta k_{p,q,r} = \sum_{n=1}^{N} f_n(x_{i+p,j+q,k+r}) \cdot \omega_{n,p,q,r}$$

where $f_n$ is the feature extraction function applied to the input features, $\omega_n$ is the set of learnable weights associated with the $n^{th}$ feature channel, and $N$ is the number of feature channels. RD3C allows the convolution kernel to be adaptively adjusted to the input features, which can better capture the variations and deformations in 3D data, making it well-suited for object detection tasks in high-resolution remote sensing images of oil storage tanks.

### B. 3.2-Comparing YOLOv5 and EfficientDet

Mekhalfi et al., [17] initially perform a contrastive study and provides enough evidence that proves, even though EfficientDet results higher mAP but YOLOv5 can detect more examples and has better generalization capabilities. They reproduce results on EfficientDet and list out intuitions behind using BiFPN over FPN as follows:

1. Including nodes with one input edge will have a smaller contribution in feature fusion. (Yellow nodes in Figure 2)

2. Extra edge ties the input node to the output node. (Green and blue edges from input to output nodes)

3. Each bidirectional path is considered as one feature layer, repeated several times to enable high-level feature fusion. (Up down arrows in Figure 2)

### C. 3.3-Automated Defect Detection: Modifying Backbone

Even though Medak et al., in [18] agree that object detection algorithms require large amount of data to provide human-level accuracy, they prove EfficientDet to be able to perform SOTA results on realistic performance in Ultrasonic and Forensics defect detection. They introduce a novel anchors (sliding window) size finding mechanism for OSOD, a kind of hyperparameter search. Anchors are predefined rectangles used by one-stage detectors to predict object locations and sizes. In this case, the hyperparameters are calculated using a novel procedure that considers the aspect ratio of the defects in UT images. This improves the detection of defects with extreme aspect ratios and increases the model's average precision. The complete novelty of this approach can be explained with the following Algorithm 1. It involves K-means clustering with Jaccard distance to calculate new values for aspect ratios and scales, and finding the template anchor size that is most like the calculated shape to determine the scale factor. The final values greatly differ from commonly used default values and were found to improve the performance of the EfficientDet model in detecting defects.

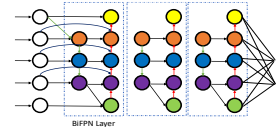

*Figure 2* BiFPN Feature-Fusion (Bottleneck of original EfficientDet)

```
Algorithm 1  Calculation of EfficientDet anchors' hyperparameters
Require: List of bounding boxes B₁, B₂, ..., Bₙ
Ensure: List of aspect ratios and scales for EfficientDet anchors
 1: Calculate the k shapes of anchors using K-means with the Jaccard distance
    metric on the list of bounding boxes:
 2:      IOU(Bᵢ, Aⱼ) = area(Bᵢ ∩ Aⱼ)/area(Bᵢ ∪ Aⱼ)
 3:      Dᵢⱼ = 1 − area(Bᵢ ∩ Aⱼ)/area(Bᵢ ∪ Aⱼ)
 4:      A₁, A₂, ..., Aₖ ← K-means(B₁, B₂, ..., Bₙ, D, k)
 5: Calculate aspect ratios:
 6:      aspectratioⱼ = height(Aⱼ)/width(Aⱼ)
 7: Calculate scales:
 8:      sᵢ = max(width(Bᵢ), height(Bᵢ))
 9:      T_{Aᵢ} = arg min |size(Tⱼ) − sᵢ|, where Tⱼ ∈ 32, 64, 128, 256, 512
10:      scaleᵢ = max(width(Aᵢ), height(Aᵢ))/size(T_{Aᵢ})
11: Merge similar scales:
12:      S = scale₁, scale₂, ..., scaleₖ
13:      Sort S in increasing order           j ← 1
14:      while j < |S| :
16:          q ← j + 1
17:          while q ≤ |S| and S_q/S_j < merge_threshold :
18:              q ← q + 1
19:          scale_j ← (∑_{i=j}^{q-1} Sᵢ)/(q − j)
20:          j ← q
21: return                       aspectratio₁, aspectratio₂, ..., aspectratioₖ,
    scale₁, scale₂, ..., scaleₖ
```

*Algorithm 1* This algorithm calculates the aspect ratios and scales for EfficientDet anchors using K-means with the Jaccard distance metric on a list of bounding boxes, and merges similar scales to produce the final list of hyperparameters.

### D. 3.4-Multilayer 3D Attention Mechanism

The combination of feature fusion with multilayer attention helps to extract features from low-level visibility keeping feature channel intact for multi-scale inputs. This research work [19] proposed a method for classifying military ships from high-resolution optical remote sensing images using a multilayer feature extraction network inspired by EfficientDet trackers. In the proposed method, a multilevel attention mechanism was used to effectively extract multilayer features, and a deep feature fusion network was constructed to locate and distinguish different types of ships. In contrast, our approach for marine animal and species detection uses a modified EfficientDet network with skip connections to improve accuracy, rather than using the proposed method. Residual connections are a type of skip connection used in deep neural networks, but they have some limitations compared to standard skip connections.

A simple architecture of both where a simple XOR is performed for input and output of previous layer in residual connection, however a connection is broken in skip connection which is shown by dashed line in Figure 3.

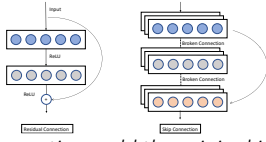

*Figure 3* Residual connections add the original input to the output of the previous layer, while skip connections concatenate the input and output of the previous layer.

Residual connections can help with the vanishing gradient problem and improve the performance of deep neural networks, but they are not as effective as skip connections in some cases. Skip connections are more flexible and can learn complex representations better than residual connections, especially when the number of layers is large. In general, the choice of which to use depends on the specific problem being addressed, therefor for our application we use skip connections.

### E. 3.5-Another FashionNet

In the field of fashion image analysis, deep learning models with high computational requirements have been a challenge. To address this, the proposed study [20] presents a one-stage detector that can rapidly detect multiple clothes and landmarks in fashion images. The study modifies the EfficientDet network and applies compound scaling to the backbone feature network. The bounding box/class/ landmark prediction network maintain the balance between the speed and accuracy. With an inference time of 42 ms and without image preprocessing, the proposed model achieves a mean average precision (mAP) of 0.686 in bounding box detection and 0.450 mAP in landmark estimation on the DeepFashion2 validation dataset, making it efficient for real-world applications. The research also highlights the potential for training networks on specific domains for computer vision applications and future fashion image analysis research. While the proposed network achieved a balance between speed and accuracy, it still had high computational requirements and could not perform low-level feature learning effectively. In contrast, my proposed algorithm improves upon this by modifying the EfficientDet network with a modified BiFPN bottleneck with skip connections. This modification makes the network more robust and enables it to learn low-level features without forgetting and gradient loss. This improvement can lead to better detection and classification of marine animals and species, as well as more efficient use of computational resources. By incorporating skip connections, my approach ensures that the low-level features are preserved throughout the network, resulting in better accuracy and faster inference times. Details and working procedure of their work is summarized with Algorithm 2.

Summary of the related work surveyed, detailed in Table 1.

### IV. PROPOSED METHODOLOGY

A recent report by MarketsandMarkets [21] has projected that the global maritime analytics market size will reach USD 7.8 billion by 2025, exhibiting a Compound Annual Growth Rate (CAGR) of 10.0% during the forecast period, up from USD 4.9 billion in 2020. This growth can be attributed to the increasing adoption of advanced technologies like artificial intelligence, machine learning, and big data analytics in the maritime industry. In this context and its rising importance, I propose a robust mechanism that strikes a balance between speed, accuracy, and inference time for real-time performance, utilizing a modified EfficientDet model. In line with the relevant literature, it has been established that EfficientDet is a state-of-the-art algorithm for performing complex tasks in the maritime, military, and remote sensing domains.

---

**Algorithm 2** Fashion Landmark Detection

1: **Input:** Fashion image dataset
2: **Output:** Fashion landmarks and bounding boxes
3: **Model:** EfficientDet with modifications
4:   - Backbone: EfficientNet
5:   - Feature Pyramid Network: BiFPN
6:   - Prediction Head: 3x3 convolution followed by convolution with filter size 3, stride 1, and padding 1
7:   - Number of anchors: 9
8:   - Number of classes: 13
9:   - Number of landmarks: 294
10:  - Loss Function: Focal Loss, Complete IoU Loss, Rooted Mean Squared Error
11:
12: **Procedure:**
13:  1. Pretrain EfficientNet backbone
14:  2. Modify EfficientDet with BiFPN structure
15:  3. Design prediction head for fashion landmarks and bounding boxes
16:  4. Train the model using Focal Loss, Complete IoU Loss, Rooted Mean Squared Error
17:
18: **Loss Function:**
19:  - Focal Loss for classification:
20:    $L_{cls} = -\alpha_t(1-p_t)^\gamma \log(p_t)$
21:    $\alpha = 0.25, \gamma = 2$
22:
23:  - Complete IoU Loss for bounding box regression:
24:    $V = \pi^2(\arctan(h) - \arctan(h_{gt}))$
25:    $\alpha = V$ if IoU $< 0.5$, else $\alpha = 1 - IoU + V$
26:    $L_{bbox} = 1 - IoU + c_w^2 + c_h^2 + \alpha V$
27:
28:  - Rooted Mean Squared Error for landmark prediction:
29:    $v$ is the visibility of clothing landmarks. If $v$ does not exist, it is not reflected in the loss.
30:    $L_{landmark} = \sqrt{\frac{1}{n}\sum_{i=1}^{n}(y_i - \bar{y}i)^2}$ if $v > 0$, else $Llandmark = 0$
31:
32:  - Total Loss:
33:    $L_{tot} = L_{cls} + L_{bbox} + \lambda_{size}L_{landmark} + \lambda_{off}L_{off}$
34:    $\lambda_{size} = 0.1, \lambda_{off} = 1$

*Algorithm 2* Algorithm for Landmark Estimation using a modified version of EfficientDet and BiFPN structure to balance speed and accuracy, with a prediction network designed specifically for fashion images and four loss functions formulated for object detection.

Object detection models typically consist of three main components - the backbone, neck, and head. These components are responsible for feature extraction, spatial feature fusion, and object detection, respectively. In this work, we devise a modified version of the EfficientDet object detection model. To this end, I propose a robust bottleneck layer, BiSkFPN, which is an abbreviated form of Bidirectional Skip-Connection-based FPN, that can learn murky inputs feature maps. Further, I improve accuracy using adversarial learning and increase the interpretability of the black-box object detection model. Finally, I use the GradCAM++ [25] algorithm to visualize the class activation maps in the last sub-section.

**A. Dataset:** The original Brackish Dataset [15] contains captured videos from the source in the AVI format for each of the six classes. However, authors have published dataset for five classes on Kaggle [22]. Further, as shown in [15], using *ffmpeg* library, the videos were converted to frames. The dataset was downloaded from Kaggle and stored to google drive space. Size of original dataset is 174 MB in AVI format (14518 frames). Each video file is then converted to PNG images using *ffmpeg* with scaling of 960x540. This width and height were chosen by authors, so I also had to choose same to match annotations coordinates. Further to verify the images and their corresponding annotations co-exist, I wrote a script that compares the files list from *"images"* and *"annots"* folder. After that I use imagekick-mogrify library to convert PNG to JPG. The final step was to normalize annotation coordinates. The purpose of this function is to normalize the bounding box coordinates in the YOLO text file to values between 0 and 1, relative to the dimensions of the original image. The function first sets the dimensions of the original image to 960x540. It then normalizes these coordinates by dividing each value by the width or height of the image. Normalizing coordinates in object detection tasks, like YOLO, is important as it helps make the model more robust and generalizable by scaling the coordinates to a common range (0 to 1) across different image sizes and preventing the model from being biased towards certain aspect ratios or image sizes. After finishing this process, a 70:20:10 ratio is maintained over all experiments for train-val-test dataset. Also, I use RoboFlow [23] to host my pre-processed dataset online with Rotations, Orientation, Flipping, Cropping and Scaling augmentations for robustness. Further details of the dataset are summarized in Table 2.

The standard deviation is a measure of the spread or variability of a dataset. In this case, the standard deviation of approximately 7.47 indicates that the number of files in each category is relatively spread out or variable, with some classes having many files and others having relatively few. This information can be useful in understanding the class-wise distribution of files in a dataset and identifying any potential imbalances or biases. For example, the "fish-big" category has 29 files, which is more than one standard deviation above the mean, while the "shrimp" category has only 8 files, which is more than one standard deviation below the mean. All extra frames and images that do not have annotations were removed, thus discarding approximately 3700 images without labels and 1807 images with no visibility, and finally resulting 10000 total images with 10000 corresponding image annotations. Annotations are allowed to have more than one class object per example. For example, in Figure 4, three example images have multiple bounding boxes representing multiple animals (objects). Finally, the class wise distribution of images in the updated dataset is shown in Table 3 and Figure 5.

**B. Proposed Model Architecture:** The EfficientDet model [7] is developed on top of three components of Single Stage Object Detection namely: Backbone (EfficientNet), Bottleneck (BiFPN) and Head (YOLOv3 like). I intend to modify the backbone with swish activation function, bottleneck proposing BiSkFPN and head using Multi Focal Loss, keeping the backbone intact. In this section I give detail explanation of the three.

**B.1. Backbone:** Spatio-Resoluion architecture from EffNet [11] performs a volumetric 3D convolution over input images which enhances the depth-width-resolution based features with upscaling and downscaling to learn every possible feature maps. This is helpful because marine objects come with varying shapes, sizes, angle of captured image, resolution, and noise. I introduce mobile inverted bottleneck convolution (MBConv) block that uses depth wise separable convolutions to reduce the number of parameters and improve the computational efficiency of the network. The MBConv block also includes a shortcut connection that allows the network to learn more complex features and deeper representations. Further, I incorporate a new type of normalization called swish activation, which has been shown to outperform traditional *ReLU* activation functions. Swish activation is a smooth non-linear function that is easy to compute and leads to better generalization performance. EfficientNet is designed to handle a wide range of input sizes and aspect ratios. The architecture includes a series of convolutional layers and pooling operations that down sample image to generate a feature map with a fixed resolution. These layers have been kept intact from the standard model. The resolution of the feature map is determined by the scaling factor $\alpha$, and the size of the input image. EfficientNet introduces the MBConv block, which is defined as, for a given input feature map $X$ with dimensions $H \times W \times C$, the MBConv block applies the following operations:

1. Depthwise Convolution: The input feature map $X$ is convolved with a depthwise convolutional filter with kernel size $K \times D \times D$, where $K$ is the number of channels, $D$ is the spatial dimension, and the output feature map is $X_d$ with dimensions $H \times W \times K$.

2. Pointwise Convolution: The output feature map $X_d$ is then convolved with a pointwise convolutional filter with kernel size $1 \times 1 \times (K \times T)$, where $T$ is the expansion factor. The output feature map is then $X_p$ with dimensions $H \times W \times (K \times T)$.

| # | Goals | Methodology | Results | Advantages | Disadvantages |
|---|-------|-------------|---------|------------|---------------|
| 1 | Oil tank detection using remote sensing imagery [16] | Modification of the model Training EffDet | Higher accuracy and less inference time (100 mAP) | Suitable for IoT based Obj Detection | Suitable for remote sensing, not marine vision |
| 2 | Compare OD models for crop circle detection in desert [17] | Annotated dataset to train YOLO and Efficient Det | EfficientDet achieves higher accuracy than other models (91 mAP) | Useful comparison based on feature extraction layers of different OD models | Overfitting, Complexity, Memory requirements, difficult to fine tune for specific task. |
| 3 | Defect detection in ultrasonic images [18] | Using bi-level CNN and vanilla NN for task | Beats SOTA with high accuracy (89.65 mAP) | Cheap localization, simple regression loss | Not uses any standard object detection baseline |
| 4 | Detecting military ships in high-resolution optical remote sensing images [19] | Using pretrained backbone, FPN layer for feature extraction (at different scales) and regular NN for class loss for annotated dataset. | EfficientDet achieves higher accuracy in remote sensing imagery (97.05 mAP) | The technique and application has potential to extend to underwater marine vision | The model still uses old FPN layer which is prone to catastrophic forgetting, and label dispersion |
| 5 | Defect detection and landmark for Clothing Dataset [20] | Using SSOD for annotated clothing dataset | Finds fashion landmark (apparel factor) in the clothing (68.60 mAP) | Uses curriculum learning, along with SSOD, better generalization | Take 4 times more time to train and could lead to large parameter size ~B |

*Table 1 Summary of related work and its key details*

3. **Swish Activation:** The output feature map $X_p$ is then passed through the swish activation function defined as $\text{Swish}(x) = x \times \text{sigmoid}(x)$.
4. **Projection:** The output feature map $X_p$ is then convolved with a pointwise convolutional filter with kernel size $1 \times 1 \times T$, where $T$ is the reduction factor. The output feature map is then $X_{\text{out}}$ with dimensions $H \times W \times T$.
5. **Shortcut Connection:** The input feature map $X$ is added elementwise to the output feature map $X_{\text{out}}$, and the resulting feature map is then passed through a ReLU activation function.

| Property | Value |
|---|---|
| Image Format | JPG |
| Resolution | 960X540 |
| Total Images | 10000 |
| Train | 7000 |
| Test | 1000 |
| Val | 2000 |
| Annotation Format | YOLO-TXT, COCO-JSON |
| Classes | 0-5 (Crab, Fish-Big, Fish-school, fish-small, shrimp, jellyfish) |
| Class wise distribution of AVI files | 17, 29, 9, 22, 8, 12 |
| Mean | 16.17 |
| Variance | 55.80 |
| Standard Deviation | 7.47 |

*Table 2 Details of the dataset with class distribution of video files*

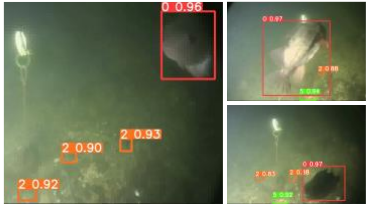

*Figure 4 Three example images from training dataset with classes "Fish-big", "Crab" and "Jellyfish"*

| Class | Number of Examples |
|---|---|
| Crab | 1751 |
| Fish-big | 2992 |
| Fish-school | 927 |
| Fish-small | 2268 |
| Shrimp | 824 |
| Jellyfish | 1237 |

*Table 3 Class wise breakage of image examples in the dataset*

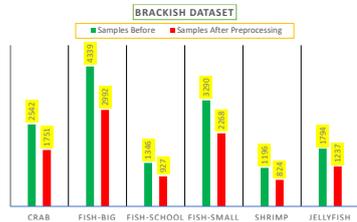

*Figure 5 Number of statistics before and after data preprocessing*

The swish activation function shown in Figure 6 is defined as:

$$\text{Swish}(x) = x \times \text{sigmoid}(\beta x)$$

where β is a trainable parameter. The derivative of the swish activation function is given by:

$$\frac{d(\text{Swish}(x))}{dx} = \sigma(\beta x) + x \times \beta \times \sigma(\beta x) \times (1 - \sigma(\beta x))$$

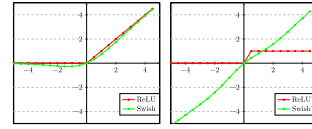

*Figure 6 ReLU vs Swish activation functions (Left) and Derivative of ReLU and Swish activation functions (Right)*

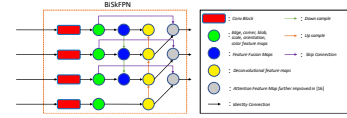

*Figure 7 Modified and proposed bottleneck to enhance feature maps with skip connection. This modification avoids catastrophic forgetting and robust against adversarial noise.*

The resolution of the feature map is determined by the scaling factor, phi, and the size of the input image. Specifically, the input image is resized to have a shorter side of size S, and the resolution of the feature map is given by: $resolution = \lceil \alpha \times S / R \rceil$ where α is scaling factor, and $R$ is the down-sampling rate of the network.

**B.2. Bottleneck:** A modified Bidirectional Skip-Connection FPN (BiSkFPN) bottleneck is proposed by modifying original BiFPN layer. BiSkFPN architecture is shown in Figure 7 which has an additional layer of *Deconv* feature maps and skip connections. On contrary to the previous research, that has used attention mechanism to preserve low-level features, I use *Deconv* feature maps which are less computationally expensive than attention mechanism and robust to perturbations.

A feature map is the output of a convolutional layer that highlights specific features in the input data. Each feature map contains a set of values that represent the presence or absence of a specific feature in the input. The purpose of the feature map is to extract high-level features from the input data that are relevant to the task at hand. Attention mechanisms are used to selectively focus on certain parts of the input data that are most relevant to the task at hand. An attention layer can be added between two convolutional blocks to learn feature weights based on the input data. Deconvolution or transposed convolution layers can be used to up-sample the feature maps and increase their spatial resolution. A deconvolution layer can be added between two convolutional blocks to recover lost spatial information.

Overall, my proposed BiSkFPN that utilizes skip connections has been shown to outperform the standard BiFPN. The skip connections allow the network to preserve low-level features and enhance the flow of information between convolutional blocks, resulting in more accurate predictions. Moreover, adding a deconvolution block after feature fusion maps in the BiSkFPN layer helps the attention mechanism to perform faster by reducing the spatial dimensions of the feature maps. This makes it easier for the attention mechanism to selectively focus on the most relevant parts of the input data (and works as a filter). Additionally, the skip connections are helping to pass information from the convolutional blocks to the attention feature maps by skipping the deconvolution feature maps. This ensures that the high-level features learned in the convolutional blocks are preserved and utilized in the final predictions, improving the accuracy and robustness of the network. The modified Bi-Scale FPN (BiSkFPN) can be expressed as follows:

$$\text{BiSkFPN}(P) = \text{concat}\left(P_i, \text{deconv}(P_{i+1}), \text{skip}_{i \to i+1}(P_{i-1})\right)$$

where $P_i$ represents the feature maps obtained after the $i^{th}$ convolutional block, $\text{skip}_{i \to i+1}$ is the skip connection between the $i^{th}$ and $(i+1)^{th}$ convolutional blocks, and deconv is the deconvolution block.

**B.3. Head:** Head performs three computations to Regularize, Classify and Regress. The head in EfficientDet architecture consists of two sub-networks, one for classification and another for regression. The classification network predicts the class probabilities for each anchor box, and the regression network predicts the box offsets for each anchor box. The classification loss is a cross-entropy loss that measures the difference between the predicted class probabilities and the ground truth class labels. The cross-entropy loss is defined as follows:

$$L_{cls} = -\frac{1}{N}\sum_{i=1}^{N}\sum_{c=1}^{C} y_{i,c} \log(\widehat{y_{i,c}})$$

where N is the number of anchor boxes, C is the number of classes, $y_{i,c}$ is the ground truth class label for the $i$-th anchor box and $c$-th class, and $\widehat{y_{i,c}}$ is the predicted probability of the $i$-th anchor box belonging to the $c$-th class. The regression loss is an $L1$ loss that measures the difference between the predicted box offsets and the ground truth box coordinates. The $L1$ loss is defined as follows:

$$L_{box} = \frac{1}{N}\sum_{i=1}^{N}\sum_{j \in x,y,w,h} \text{smooth}_{L1}(t_j - \hat{t}_j)$$

where $N$ is the number of anchor boxes, $t$ is the ground truth box coordinates, $\hat{t}$ is the predicted box coordinates, and $\text{smooth}_{L1}$ is the smooth $L1$ loss defined as:

$$\text{smooth}_{L1}(x) = \begin{cases} 0.5x^2, & \text{if } |x| < 1 \\ |x| - 0.5, & \text{otherwise} \end{cases}$$

The regularization loss is an $L2$ norm that measures the magnitude of the network weights. The $L2$ regularization loss is defined as follows:

$$L_{reg} = \frac{1}{N}\sum_{i=1}^{N} |w_i|_2^2$$

where $N$ is the number of network weights and $w_i$ is the $i$-th weight. The total loss function is a weighted sum of the classification loss, regression loss, and regularization loss, defined as follows:

$$L = L_{cls} + \alpha L_{box} + \beta L_{reg}$$

where α and β are hyperparameters that control the relative importance of the different loss terms. The working algorithm for the proposed EfficientNet backbone is provided in Algorithm 3, for the proposed BiSkFPN in Algorithm 4 and the modified MFL head in Algorithm 5.

**B.4. Universal Adversarial Perturbation (UAP):** UAP is a type of noise that can be added to any input image to fool a deep neural network (DNN) into misclassifying it. It is computed as the sum of a small random perturbation vector $\delta$ and a direction vector $r$, both in the image space, and scaled by a constant $\xi$. Formally, the UAP is given by:

$$U = \xi \cdot \text{sign}\left(\sum_{i=1}^{n} r_i \cdot \delta_i\right)$$

where $\delta_i$ and $r_i$ are the $i$-th elements of the perturbation and direction vectors, respectively, and $n$ is the total number of pixels in the image [24]. Adding UAP to the marine vision dataset can make it more robust to adversarial attacks during training via adversarial learning. This is because training the network with both clean and perturbed images can help it learn to be more robust to perturbations and generalize better to new, unseen data. This can be particularly important for applications such as aquatic life detection, where robustness to variations in lighting, water quality, and other environmental factors is crucial for accurate detection and classification.

---
**Algorithm 3** EfficientDet Backbone Algorithm
**Require:** Input Image $I$, Scaling factor $\alpha$, Down-sampling rate $R$, Depthwise convolution kernel size $D$, Expansion factor $T$, Reduction factor $T'$
**Ensure:** Feature map with fixed resolution
1: **function** EFFICIENTDETBACKBONE($I, \alpha, R, D, T, T'$)
2:   $S \leftarrow$ minimum dimension of $I$
3:   $resolution \leftarrow \lceil \alpha \times S / R \rceil$
4:   $X \leftarrow$ input feature map of size $H \times W \times C$
5:   $K \leftarrow$ number of channels in $X$
6:   $X_d \leftarrow$ depthwise convolution on $X$ with kernel size $K \times D \times D$:
7:   $\forall i, j, k : X_d(i, j, k) = \sum_{u,v} X(i + u, j + v, k) \times W_d(u, v, k)$
8:   where $W_d$ is the depthwise convolution kernel of size $D \times D \times K$
9:   $K' \leftarrow K \times T$
10:   $X_p \leftarrow$ pointwise convolution on $X_d$ with kernel size $1 \times 1 \times K'$:
11:   $\forall i, j, k' : X_p(i, j, k') = \sum_k X_d(i, j, k) \times W_p(1, 1, k, k')$
12:   where $W_p$ is the pointwise convolution kernel of size $1 \times 1 \times K \times K'$
13:   $X_p \leftarrow$ swish activation function on $X_p$ with trainable parameter $\beta$:
14:   $\forall i, j, k' : X_p(i, j, k') = \frac{X_p(i,j,k')}{1+\exp(-(\beta \times X_p(i,j,k')))}$
15:   $X_{\text{out}} \leftarrow$ pointwise convolution on $X_p$ with kernel size $1 \times 1 \times T'$:
16:   $\forall i, j, k : X_{\text{out}}(i, j, k) = \sum_{k'} X_p(i, j, k') \times W_{\text{out}}(1, 1, k', k)$
17:   where $W_{\text{out}}$ is the pointwise convolution kernel of size $1 \times 1 \times K' \times T'$
18:   $X_{\text{out}} \leftarrow$ elementwise addition of $X$ and $X_{\text{out}}$:
19:   $\forall i, j, k : X_{\text{out}}(i, j, k) = X(i, j, k) + X_{\text{out}}(i, j, k)$
20:   $X_{\text{out}} \leftarrow$ ReLU activation function on $X_{\text{out}}$:
21:   $\forall i, j, k : X_{\text{out}}(i, j, k) = \max(0, X_{\text{out}}(i, j, k))$
22:   **return** feature map with resolution $resolution$
23: **end function**

---
*Algorithm 3 EfficientDet-Backbone: A streamlined algorithm for generating feature maps with fixed resolution from input images using depthwise and pointwise convolutions with swish and ReLU activation functions.*

---
**Algorithm 4** Bottleneck BiSkFPN Algorithm
**Require:** Input feature maps $\mathbf{P} = \mathbf{P}1, \mathbf{P}2, ..., \mathbf{P}n$, Deconvolution kernel size $K$, Skip connection feature map **skip**
**Ensure:** Fused feature map **F**
1: **function** BOTTLENECKBISKFPN($\mathbf{P}, K, \text{skip}$)
2:   $\mathbf{P}' \leftarrow$ upsample $\mathbf{P}n$ by deconvolution with kernel size $K$
3:   $\mathbf{F} \leftarrow \text{concat}(\mathbf{P}n, \mathbf{P}')$
4:   **for** $i \leftarrow n - 1$ to $1$ **do**
5:     $\mathbf{P}' \leftarrow$ upsample $\mathbf{F}$ by deconvolution with kernel size $K$
6:     $\mathbf{F} \leftarrow \text{concat}(\mathbf{P}i, \mathbf{P}')$
7:     $\mathbf{F} \leftarrow \mathbf{F} + \text{skip}(\mathbf{P}_{i-1})$
8:   **end for**
9:   **return F**
10: **end function**

---
*Algorithm 4 Bottleneck-BiSkFPN: A fusion algorithm that combines input feature maps with skip connections using deconvolution and concatenation for generating a final fused feature map.*

**B.5. GradCAM++:** Gradient-weighted Class Activation Mapping++ [25] is an explainable artificial intelligence (XAI) technique that provides visual explanations of a neural network's decision-making process. It highlights the regions in an image that are most important for the network's classification decision. The method computes the class-specific localization maps of the input image by weighing the activations of the final convolutional layer with the gradients of the output class score with respect to the activations. The

final map is obtained by summing the weighted maps. This process can be expressed mathematically as follows:

$$L^c_{GradCAM++}(x,y) = ReLU\left(\sum_k^K \alpha_k^c A_k(x,y)\right)$$

where $L^c_{GradCAM++}(x,y)$ is the GradCAM++ heatmap for class $c$, $A_k(x,y)$ is the activation of the $k^{th}$ feature map of the final convolutional layer at spatial location $(x,y)$, and $\alpha_k^c$ is the weight assigned to the $k^{th}$ feature map for class $c$. This technique is helpful in understating the specific feature distribution that model has learned for deep-sea objects in marine biodiversity.

```
Algorithm 5 Object Detection Head
Require: Anchor boxes, ground truth labels and box coordinates, hyperpa-
    rameters α and β
Ensure: Predicted class probabilities and box offsets
 1: function EFFICIENTDETHEAD(AnchorBoxes, Labels, BoxCoordinates,
    α, β)
 2:    N ← number of anchor boxes
 3:    C ← number of classes
 4:    y_{i,c} ← ground truth class label for the i-th anchor box and c-th class
 5:    ŷi,c ← predicted probability of the i-th anchor box belonging to the c-th
    class
 6:    t ← ground truth box coordinates
 7:    t̂ ← predicted box coordinates
 8:    w_i ← i-th weight of the network
 9:    Lcls ← −(1/N) Σ_{i=1}^N Σ_{c=1}^C y_{i,c} log(ŷi,c)
10:    Lbox ← (1/N) Σ_{i=1}^N Σ_{j∈x,y,w,h} smoothL1(t_j − t̂j)
11:    Lreg ← (1/N) Σ i = 1^N |w_i|^2
12:    L ← Lcls + αLbox + βLreg
13:    return ŷi,c, t̂
14: end function
```

**Algorithm 5** *EfficientDet-Head: An algorithm for predicting class probabilities and box offsets for object detection using anchor boxes, ground truth labels and box coordinates.*

## V. EXPERIMENTAL SETUP

System configuration and resources used while training different object detection model and Hyperparameter values are provided in Table 4.

| Configuration | Value | Hyperparameter | Value (YOLOv5) | Value (EffDet) | Value (DT2) |
|---|---|---|---|---|---|
| Environment | Linux (Ubuntu-Like) | Epochs | 350 | 350 | 350 |
| | | Classes | 6 | 6 | 6 |
| Service Provider | Amazon AWS (EC2) | Backbone | CSP Darknet 53 | EfficientNet | ResNet |
| | | Bottleneck | PANet | BiFPN | RPN+FPN |
| Instance Family | p3dn | Head | Yolov3 like | Yolov3 like | RPN+RCNN |
| | | Train Data | 7000 | 7000 | 7000 |
| vCPU | 96 | Val Data | 2000 | 2000 | 2000 |
| | | Test Data | 1000 | 1000 | 1000 |
| Memory | 76 GiB | Annotation/Mask format | YOLO TXT | COCO JSON | COCO JSON |
| | | Optimizer | SGD | SGD + Adam | SGD + Adam |
| Cost (per hour) | 31.2 USD | Learning rate | 0.1 | Adaptive | Adaptive |
| GPU | NVIDIA V100 TensorCore | Weight Decay (prevent overfit) | 0.0005 (Default) | 0.0005 (Default) | 0.0005 (Default) |
| | | Activation | Leaky ReLU | Leaky ReLU | Sigmoid (Bbox) + Softmax (Class) |

**Table 4** *System Configuration used (Left) and Hyper-parameter values for YOLO, EfficientDet and Detectron2 training procedure (Right). However, EfficientDet also combines Swish activation (bottleneck) along with LeakyReLU (head). Weight decay is a regularization technique that adds a penalty term to the loss function during training to encourage the model's weights to be smaller, thereby reducing overfitting.*

The object detection models used are YOLOv5, EfficientDet, and Detectron-2. For other variants of YOLO, similar configuration was used that was used in YOLOv5. A five-fold experiment is performed and averaged for results. First, I reproduce results on YOLOv3 and YOLOv4 as shown in [15]. Then, I use YOLOv5, YOLOv8, modified EfficientDet (proposed in this research) and Detectron2 to compare results. Images are resized to 416x416 by YOLO (all versions) and Detecton2 object detection models as input while original size 960x540 is used for EfficientDet. High Quality Brackish Dataset is always downloaded on the go by using RoboFlow API toolkit [23]. For the modified EfficientDet and other models' training process, batch gradient descent or stochastic gradient descent (SGD) is used with 64 batch-size. Using a moderate batch-size can lead to more stable gradient estimates and faster convergence, which can ultimately result in a higher quality model. This is especially true for larger models, where the gradient updates can be very noisy when using a smaller batch-size (for example batch size of 1). Additionally, using a larger batch size can take advantage of the parallel processing capabilities of modern GPUs, leading to faster training times. Finally, I perform adversarial training using added UAP noise to the training data in a curriculum fashion for YOLOv5 and EfficientDet models. UAP is a type of noise added to data to fool machine learning algorithms, and it is imperceptible to human senses while mitigation strategies such as adversarial training and noise-detection mechanisms are used to address this vulnerability in machine learning. Finally, GradCAM++ is used to visualize the saliency maps in latent space for class specific features which is discussed and shown in results section for both the models. A short overview of full experimentation is also shown in Figure 12.

| Model Name | mAP$_1$ | mAP$_2$ | mAP$_3$ | mAP$_4$ | mAP$_5$ | Mean±Std |
|---|---|---|---|---|---|---|
| YOLOv3 [15] | 31.9 | 30.2 | 29.5 | 32.5 | 31.7 | 31.1±1.1 |
| YOLOv4 [15, 29] | 84.6 | 83.8 | 84.2 | 85.2 | 80.9 | 83.7±1.4 |
| YOLOv5 [17] | 96.7 | 98.0 | 97.5 | 98.5 | 97.3 | 97.6±0.61 |
| YOLOv8 [26, 27] | 98.0 | 98.5 | 98.3 | 98.1 | 98.2 | 98.2±0.17 |
| Detectron2 [28] | 94.5 | 93.4 | 94.8 | 95.7 | 97.8 | 95.2±1.4 |
| Proposed EfficientDet | 99.5 | 98.7 | 98.0 | 97.0 | 99.8 | 98.6±1.0 |

**Table 5** *mAP and average mAP for five experiments with same configuration for different models on the brackish dataset (Mean ± Std). To the best of my knowledge, no previous work has utilized the brackish dataset for YOLOv5 and the models ranked below it in the table. Therefore, I replicated the same experimental setup used for these models and produced the results. The column names sub-script shows the number of experiment.*

| Model name | crab | Fish-big | Fish-school | Fish-small | Shrimp | Jellyfish |
|---|---|---|---|---|---|---|
| YOLOv3 [15] | 92.7 | 89.9 | 84.0 | 62.3 | 76.6 | 82.0 |
| YOLOv4 [15, 29] | 93.1 | 78.9 | 88.2 | 59.2 | 73.2 | 83.2 |
| YOLOv5 [17] | 81.8 | 56.3 | 80.9 | 66.9 | 69.6 | 93.3 |
| YOLOv8 [26, 27] | 82.8 | 63.2 | 85.7 | 69.5 | 65.0 | 97.4 |
| Detectron2 [28] | 28.1 | 14.5 | 8.6 | 3.8 | 26.1 | 40.6 |
| Proposed EfficientDet | 89.5 | 94.6 | 87.2 | 82.1 | 79.9 | 95.2 |

**Table 6** *Class-wise accuracy (mAP). It is important to note that "Jellyfish" is easy to learn because it does not move a lot and IoU mean does not shift a lot in training and it is orderly distributed in the dataset, however, "Fish-small" moves a lot in the frames and hence IoU average is less because it is difficult to learn increasingly complex bounding boxes and it is distributed at various places in the dataset.*

## VI. RESULTS AND DISCUSSIONS

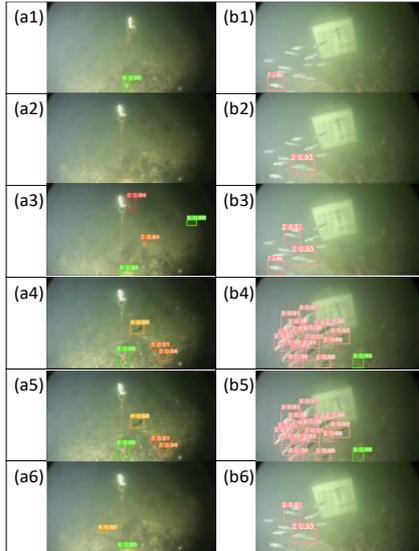

*Figure 8* Comparing all models on two test multi-class multi-label examples "a" and "b". Here (a1)-(b1) represents YOLOv3, (a2)-(b2) represents YOLOv4, (a3)-(b3) represents YOLOv5, (a4)-(b4) represents YOLOv8, (a5)-(b5) represents proposed model. (a1) to (a5) and (b1) to (b5) are one stage object detection models, while (a6) and (b6) are two stage object detection models. It is important to note that YOLOv8 and proposed EfficientDet models predict the most labels that are class specific which also justifies the Table 5

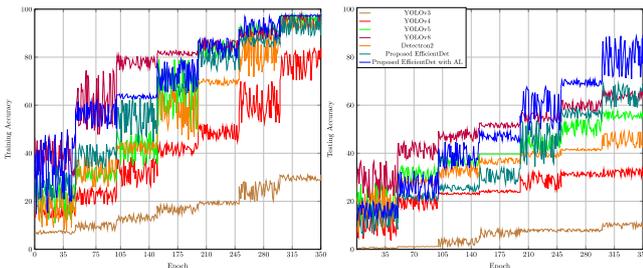

*Figure 9* Training accuracy (Left) and Testing accuracy (Right) for 350 Epochs. Results are shown for YOLOv3 [15], YOLOv4 [15, 29], YOLOv5 [17], YOLOv8 [26, 27], Detectron2 [28], Proposed EfficientDet and Proposed EfficientDet with Adversarial Learning (AL) with UAP noise. The sudden jerks in the plot show that the weight decay performing well to punish higher chances of overfitting.

To evaluate the performance of the object detection models, there are some key technical aspects related to results need to be understood before we finally see the numbers on different evaluation metrics for multi-class multi-label problem (ground class label is the original label of an example):

1. *True Positive (TP):* # of times a ground class label is predicted correctly, for example, "small-fish", is predicted as "small-fish".

2. *False Positive (FP):* # of times a ground class label is predicted incorrectly as belonging to a particular class. For example, if we are trying to classify "small-fish", FP represents the number of instances of a different species (classes may be "crab", "big-fish", "jellyfish", etc) that are incorrectly identified as the specie we are interested in (that is "small-fish").

3. *True Negative (TN):* # of times a ground class label is correctly classified as not belonging to a particular class. For example, if we are trying to classify "small-fish", TN represents the number of instances of all other species (classes may be "crab", "big-fish", "jellyfish", etc) that are correctly identified as not belonging to the specie ("small-fish") we are interested in.

4. *False Negative (FN):* # of times a ground class label is incorrectly classified as not belonging to a particular class. For example, if we are trying to classify "small-fish", FN represents the number of instances of the specie ("small-fish") that are incorrectly identified as a different species (classes may be "crab", "big-fish", "jellyfish", etc).

5. *True Positive Rate (TPR) or Recall*: The ratio of true positives that are correctly identified by the classifier. TPR is also known as recall or sensitivity. $TPR = TP/(TP + FN)$. In the context of our dataset, TPR represents the ratio of instances of a particular specie that are correctly identified as that specie. For example, Recall is number of times model classified it was "small-fish" divided by the number of times it was actually "small-fish".

6. *False Positive Rate (FPR)*: The ratio of true negatives that are incorrectly identified as positive by the classifier. $FPR = FP/(FP + TN)$. In the context of our dataset, FPR represents the ratio of instances of all other species ("crab", "big-fish", "jellyfish", etc) that are incorrectly identified as the specie we are interested in ("small-fish").

7. *True negative rate (TNR)*: The ratio of true negatives that are correctly identified by the classifier. TNR is also known as specificity. $TNR = TN/(FP + TN)$. In the context of our dataset, TNR represents the ratio of instances of all other species ("crab", "big-fish", "jellyfish", etc) that are correctly identified as not belonging to the specie we are interested in ("small-fish").

8. *False negative rate (FNR):* The ratio of true positives that are incorrectly identified as negative by the classifier. $FNR = FN/(TP + FN)$. In the context of our dataset, FNR represents the proportion of instances of the specie ("small-fish") we are interested in that are incorrectly identified as a different species ("crab", "big-fish", "jellyfish", etc).

9. *Precision:* Precision measures the number of true positives divided by the number of true positives plus false positives. It is a measure of how well the model correctly identifies positive samples. $Precision = TP/TP + FN$.

Further, Confusion Matrix talks about the true positives, true negatives, false positive, false negatives. In simple terms, Confusion matrix tells about how many times "small-fish" class is classified as "crab", other species and "small-fish" actually. And same for other classes. This is best way to evaluate a model. That is, how many times our model was having confusion with "crab" or other species, but it was "small-fish" in reality. This is where we need to create a predicted label array along the original labels to identify the parameters. Various terms are used to evaluate the performance of machine learning models. *Overall accuracy* measures the proportion of correct classifications made by the model across all classes. *Class-wise accuracy* measures the accuracy of the model on a per-class basis, which can be particularly important when dealing with imbalanced datasets. *Test loss* and *train loss* measure the error of the model during training and testing, respectively. The lower the loss, the better the model's performance. *Train accuracy* and *test accuracy* measure the proportion of correct classifications made by the model during training and testing, respectively. *Box loss* measures the error between predicted and ground-truth bounding boxes in object detection tasks. *Mean Average Precision (mAP)* measures the precision and recall of object detection models by comparing predicted bounding boxes to ground-truth boxes. *Intersection over Union (IoU)* is a measure of how well the predicted and ground-truth bounding boxes overlap. IoU is useful to evaluate how well the model can localize objects. *Class-wise accuracy* measures the accuracy of the model on a per-class basis, which can be particularly important when dealing with imbalanced datasets. Each of these terms is important in assessing the performance of our object detection models and help guiding model selection and improvement.

A five-fold experiments mAPs for the models used on the same dataset as shown in Table 5 where proposed EfficientDet and YOLOv8 have high average accuracy and acceptable range of standard deviation. Mean ± Std is a statistical measure that indicates the average performance of a model across multiple trials, along with the variability in the results. The mean represents the average value of a set of data, while the standard deviation is a measure of how much the individual data points deviate from the mean.

The class-wise accuracy for each of the model is shown in Table 6. It can clearly be seen in Table 6 that proposed model is easily able to learn difficult class features along with the box coordinates of "*Fish-small*" with higher class-mAP (82.1) while other models struggle to even reach near that score. Figure 8 is shown to compare the testing performance of the above-mentioned models and their predicted bounding boxes.

Figure 9 shows train-test accuracy curves of different models used in the experiment to compare. The train and test accuracy curve is a plot of the accuracy of a model on the training and testing datasets over epochs, as the model is trained. Furthermore, by utilizing UAP adversarial noise as part of the adversarial learning approach in the training procedure of EfficientDet for the dataset, the proposed approach demonstrates significant improvements in both training and testing accuracy and loss, also depicted in Figure 9. UAP was utilized in object detection training procedure by adding small perturbations to the input image. In addition, such noise is imperceptible to time-limited humans [30]. Therefor I provide an example in Figure 10 where the same ground truth image was tested for proposed EfficientDet model before and after UAP attack, along with the predictions from proposed EfficientDet with Adversarial Learning (AL). Same configuration was used for AL as mentioned for EfficientDet in Table 4 and methodology proposed in section 4.2.4. Finally, the GradCAM++ tool (introduced in section 4.2.5) is utilized to understand the class-specific features for the proposed EfficientDet model and YOLOv8 shown in Figure 11. It is particularly useful for understanding how the model makes its decision, and which parts of the image are most important for that decision. Visualizing class-specific features using GradCAM++ is important because it can help us understand why a model is making a certain prediction, and whether it is focusing on the correct features. By visualizing the regions of an image that are most relevant for a specific class, we can gain insights into the model's decision-making process and potentially improve its accuracy.

## VII. CONCLUSIONS

DeepSeaNet is built on top of modified EfficientDet which uses EfficientNet backbone, BiSkFPN bottleneck and multi-focal loss head. Comparative experiments with YOLOv3, YOLOv4, YOLOv5, YOLOv8, Detectron2 and proposed EfficientDet successfully prove that BiSkFPN is better alternative than FPN, PANet and BiFPN. Moreover, a high-performance computational system was used to perform several experiments for All other models and proposed model with Adversarial Learning (AL) which achieves higher accuracy than EfficientDet. AL makes the model robust against real world adversarial examples, and it becomes important for underwater "Brackish dataset". Finally, I prove the results obtained while training, using feature-map visualization that shows class-specific features for YOLOv8 and proposed EfficientDet. Even though YOLOv8 achieves near-level accuracy, it still either hallucinates or provide incorrect labels that can be seen in heat-maps of GradCAM. Therefore, after this research, it is advisable to use EfficientDet in such complex environments. A full repository of working code and experiments with results is available at https://github.com/s4nyam/efficientdet-advml.

### ACKNOWLEDGMENT

This project was completed as part of Advanced Machine Learning course at Østfold University College, Halden, Norway. The instructor for the course was Prof. Ripon. Also thanks to Perparim Mustafa for continuous support to access local HPC at HiØ.

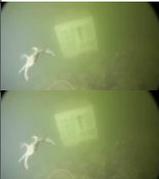

*Figure 11 In the case of "Crab" (fist two rows), it can be clearly seen that YOLOv8 is predicting a wrong object (right side box), but is not able to provide explanation for that localization, however, EfficientDet is able to predict and localize class specific features correctly with correct localization. Similarly, for "Fish-School (last two rows), YOLOv8 is not able to predict fishes due to low confidence that can be seen in feature maps, whereas EfficientDet is not only able to localize but also able to provide relevant localization feature maps with a high confidence.*

| Original Image | Ground Truth Label | Proposed EfficientDet Prediction | Prediction Label before UAP Attack | Prediction Score before UAP Attack | UAP Noise raw pixels | Ground truth Image after UAP attack | Proposed EfficientDet prediction after UAP attack | Prediction Label after UAP Attack | Prediction Score after UAP Attack | Proposed EfficientDet prediction after UAP attack with AL | Prediction Label after UAP Attack with AL | Prediction Score after UAP Attack with AL |
|---|---|---|---|---|---|---|---|---|---|---|---|---|
| 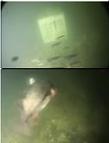 | "Fish-School" | 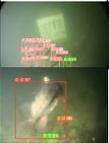 | "Fish-School" | 92.00% | 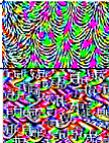 | 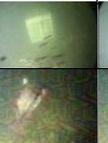 | 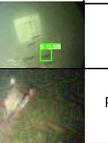 | "Crab" | 96.00% | 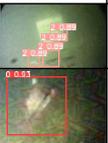 | "Fish-School" | 89.00% |
| 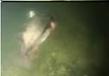 | "Fish-Big" | 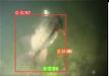 | "Fish-Big" | 97.00% | 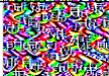 | 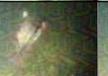 | No Prediction | NA | | 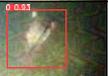 | "Fish-Big" | 93.00% |

*Figure 10* Showing two examples with their respective predictions and images before & after UAP attack and predictions after Adversarial Learning. Please note that for UAP noise in row 1, it is difficult to visualize the same in original image after attack, while that is not the case in row 2, which infers that it was difficult for "Fish-Big" image to perturb, and hence higher values of noise is added to fool the model. However, it is shown that it was easy to mitigate such adversarial noise by just making model learn those perturbed images.